\pdfoutput=1

\documentclass[11pt]{article}

\usepackage[preprint]{acl}

\usepackage{times}
\usepackage{latexsym}

\usepackage[T1]{fontenc}

\usepackage[utf8]{inputenc}

\usepackage{microtype}

\usepackage{inconsolata}

\usepackage{xspace}
\usepackage{inconsolata}
\usepackage{booktabs}
\usepackage{comment}
\usepackage{graphicx} 
\usepackage{tcolorbox} 
\usepackage{framed}
\usepackage{CJKutf8} 
\usepackage{multirow}
\usepackage{url}
\usepackage{svg}
\usepackage{soul}
\usepackage{fontawesome5}
\usepackage{xltabular} 

\usepackage{amsthm}

\usepackage{tikz}
\usepackage{forest}
\usetikzlibrary{trees,positioning,shapes,shadows,arrows.meta}
\usepackage{amssymb}
\usepackage{adjustbox}
\usepackage{rotating}
\usepackage{array}
\usepackage{makecell}
\usepackage{enumitem}
\usepackage{todonotes}

\title{Pragmatics in the Era of Large Language Models:\\ A Survey on Datasets, Evaluation, Opportunities and Challenges}

\author{
  \textbf{Bolei Ma$^\ast$\textsuperscript{1}}, 
  \textbf{Yuting Li$^\ast$\textsuperscript{2}}, 
    \textbf{Wei Zhou$^\ast$\textsuperscript{3,4}}, 
    \textbf{Ziwei Gong$^\ast$\textsuperscript{5}}, 
\textbf{Yang Janet Liu\textsuperscript{1}}, 
\textbf{Katja Jasinskaja\textsuperscript{2}}, \\
  \textbf{Annemarie Friedrich\textsuperscript{3}}, 
 \textbf{Julia Hirschberg\textsuperscript{5}}, 
  \textbf{Frauke Kreuter\textsuperscript{1,6}}, \vspace{7pt}
 \textbf{Barbara Plank\textsuperscript{1}} \\ 
  \textsuperscript{1}LMU Munich \& Munich Center for Machine Learning, 
  \textsuperscript{2}University of Cologne,
\\
\textsuperscript{3}University of Augsburg,
\textsuperscript{4}Bosch Center for Artificial Intelligence, 
  \\
    \textsuperscript{5}Columbia University, \vspace{3pt}
    \textsuperscript{6}University of Maryland, College Park
\\\vspace{2pt}
  \small{$^\ast$Equal contributions.}\\
  \small{
    \texttt{bolei.ma@lmu.de, yuting.li@uni-koeln.de, wei.zhou3@de.bosch.com, zg2272@columbia.edu}
  }
}

\begin{document}
\maketitle
\begin{abstract}
Understanding pragmatics—the use of language in context—is crucial for developing NLP systems capable of interpreting nuanced language use. Despite recent advances in language technologies, including large language models, evaluating their ability to handle pragmatic phenomena such as implicatures and references remains challenging. To advance pragmatic abilities in models, it is essential to understand current evaluation trends and identify existing limitations. In this survey, we provide a comprehensive review of resources designed for evaluating pragmatic capabilities in NLP, categorizing datasets by the pragmatic phenomena they address. We analyze task designs, data collection methods, evaluation approaches, and their relevance to real-world applications. By examining these resources in the context of modern language models, we highlight emerging trends, challenges, and gaps in existing benchmarks. Our survey aims to clarify the landscape of pragmatic evaluation and guide the development of more comprehensive and targeted benchmarks, ultimately contributing to more nuanced and context-aware NLP models.

\end{abstract}

\section{Introduction}
\label{sec:intro}

In linguistics, pragmatics studies 
how context influences the meaning of language \cite{huang2017introduction, xiang2024introduction, birner2012introduction}, and how people use language in real-life situations to convey implied meanings, emotions, and intentions. Foundational work in this field, such as Grice's (\citeyear{Grice1975}) work on implicature and the cooperative principle, Austin's (\citeyear{austin1975things}) idea of speech acts, and Sperber and Wilson's (\citeyear{sperber1986relevance}) exploration of contextual inference laid the groundwork of the study of language use. These concepts continually influence linguistic studies and also provide insights for computational methods \cite{mann1980toward, saygin2002pragmatics, hovy-yang-2021-importance, cambria2024pragmatics}. 

However, traditional Natural Language Processing (NLP) models, while competent in syntactic parsing and semantic analysis, struggle with meaning that extends beyond the literal definition of words. This gap is where pragmatics becomes essential.  
Early approaches, like rule-based systems, use explicitly coded rules for knowledge representation and match input with a knowledge base for response generation \cite{bajwa2006rule,grosan2011rule}. Later, statistical models applied probability theory to predict language behavior \cite{johnson2009statistical}, while deep learning, with transformer-based models like BERT \cite{devlin-etal-2019-bert} and GPT \cite{brown2020language}, transformed NLP by enabling the generation of contextually relevant text. 
Yet their performance in understanding pragmatics remains limited 
\cite{hu-etal-2023-fine, sileo-etal-2022-pragmatics, park-etal-2024-multiprageval}.

Recently, the emergence of Large Language Models (LLMs) has intensified the need to evaluate their human-like communication abilities, particularly for nuanced use cases requiring sophisticated pragmatic reasoning \cite[][]{hu-etal-2023-fine,Ruis2022TheGO,yerukola-etal-2024-pope}. As LLMs are increasingly deployed in real-world applications, validating their ability to understand and generate contextually appropriate and pragmatically accurate responses is crucial to ensure effective and trustworthy human-computer interactions. While recent advances in LLMs have demonstrated impressive capabilities in generating coherent and contextually appropriate text, their pragmatic competence remains insufficiently evaluated \cite[][]{hu-etal-2023-fine, kwon-etal-2023-beyond, chang2024}. 

Important questions arise: What resources have been developed to evaluate the pragmatic capabilities of NLP models, and, more importantly, how can we leverage pragmatics to guide the advancement of LLMs, formalize a comprehensive framework for evaluating pragmatics, and further advance the study of pragmatics in linguistics? 
To answer these questions, we conduct a comprehensive survey of the resources available for evaluating pragmatics in NLP: 
In \S\ref{sec:core_concepts} we introduce the core concepts in pragmatics and 
in \S\ref{sec:task_types} we review how these pragmatic phenomena have been evaluated across various NLP tasks. In \S\ref{sec:collection} we summarize how previous works build a pragmatic dataset. In \S\ref{sec:eval} we present the metrics and evaluation techniques used in current research. In 
\S\ref{sec:discussion} we highlight gaps and offer recommendations for future research.

\begin{figure*}[htbp]
\scriptsize
%
\tikzset{
    basic/.style  = {draw, text width=1.45cm, align=center, 
    rectangle, fill=green!10},
    root/.style   = {basic, rounded corners=2pt, thin, align=center, fill=gray!10, rotate=0},
    tnode/.style = {basic, thin, rounded corners=2pt, align=left, fill=pink!30, text width=10cm, anchor=center},
    xnode/.style = {basic, thin, rounded corners=2pt, align=center, fill=blue!10,text width=3cm, anchor=center}, 
    edge from parent/.style={draw=black, edge from parent fork right}
}
\begin{forest} for tree={
    grow=east,
    growth parent anchor=east,
    parent anchor=east,
    child anchor=west,
    edge path={\noexpand\path[\forestoption{edge}, ->, >={latex}] 
        (!u.parent anchor) -- +(10pt,0) |- (.child anchor) \forestoption{edge label};},
}
[Phenomena, root, l sep=11.5mm, rotate=90, child anchor=north, parent anchor=south, anchor=center, 
    [Social Pragmatics, xnode,  l sep=11.5mm,
        [\citet{ollagnier-2024-cyberagressionado, sap-etal-2019-social, sap-etal-2020-social, shaikh-etal-2023-modeling, zhang-etal-2018-personalizing, welleck-etal-2019-dialogue, yang-etal-2021-predicting,buechel-hahn-2017-emobank,oraby-etal-2016-creating}, tnode] 
        ]
    [Discourse and Coherence, xnode,  l sep=11.5mm,
        [\citet{miltsakaki-etal-2004-penn, prasad-etal-2008-penn, prasad-etal-2018-discourse, pandia-etal-2021-pragmatic,sadlier-brown-etal-2024-useful,westera-etal-2020-ted, zeyrek-etal-2018-multilingual, asher-etal-2016-discourse, reinig-etal-2024-politics,lai-tetreault-2018-discourse,wu-etal-2023-multi-task, mavridou-etal-2015-linking,miao-etal-2024-discursive,durmus-etal-2019-role}, tnode] 
        ]
    [Speech Acts and Intent Recognition, xnode,  l sep=11.5mm,
        [\citet{li2023diplomat, reinig-etal-2024-politics, braga-etal-2006-progmatica, khani-etal-2018-planning, shaikh-etal-2023-modeling, ollagnier-2024-cyberagressionado, zhang-etal-2018-personalizing, welleck-etal-2019-dialogue,Godfrey1992,rashkin-etal-2019-towards,Greco2023}, tnode] 
        ]
    [Implicature and Presupposition, xnode,  l sep=11.5mm,
        [\citet{qi-etal-2023-pragmaticqa, hu-etal-2023-fine, halat-atlamaz-2024-implicatr, nizamani-etal-2024-siga, zheng-etal-2021-grice, sap-etal-2020-social,jeretic-etal-2020-natural, kameswari-etal-2020-enhancing, koyano-etal-2022-annotating, srikanth-etal-2024-pregnant, kimyeeun-etal-2024-developing,louis-etal-2020-id,damgaard-etal-2021-ill,muller-plank-2024-indirectqa,Cong2024,sravanthi-etal-2024-pub,George2019ConversationalII,pedinotti-etal-2022-pragmatic}, tnode] 
        ]
    [Context and Deixis, xnode,  l sep=11.5mm,
        [\citet{min-etal-2020-ambigqa,sravanthi-etal-2024-pub, li2023diplomat, qi-etal-2023-pragmaticqa, sileo-etal-2022-pragmatics, hu-etal-2023-fine, shaikh-etal-2023-modeling, park-etal-2024-multiprageval, park2024pragmatic, monroe-etal-2017-colors, monroe-etal-2018-generating, westera-etal-2020-ted, zeyrek-etal-2018-multilingual,guesswhat_game,tsvilodub-franke-2023-evaluating,bao-etal-2022-learning,van-miltenburg-etal-2016-pragmatic,takmaz-etal-2020-refer,takmaz-etal-2023-speaking}, tnode]
        ]
        ]
\end{forest}
\caption{A taxonomy of pragmatic phenomena.}
\label{fig:pheno}
\end{figure*}
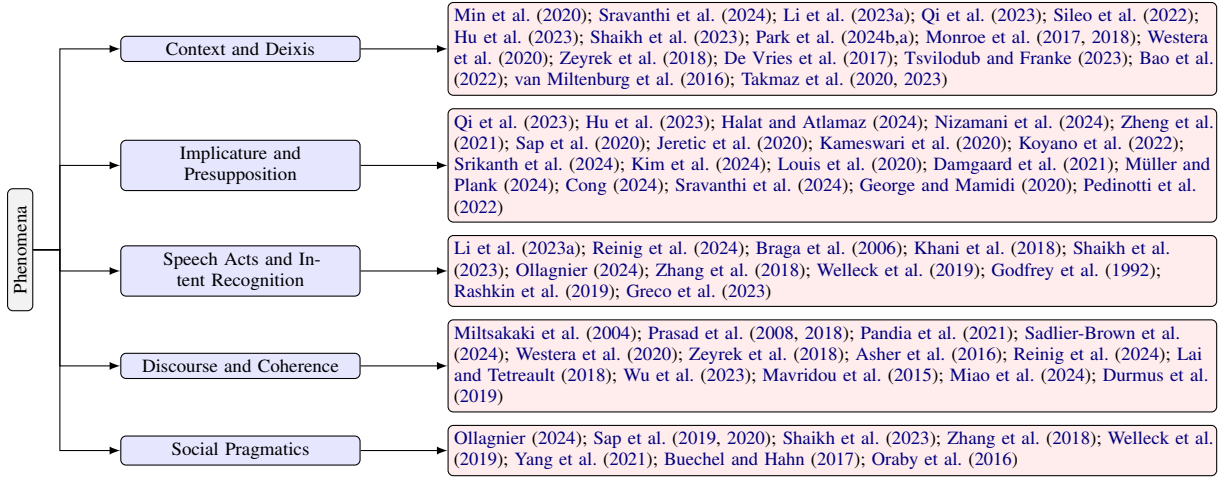

\section{Core Concepts in Pragmatics}
\label{sec:core_concepts}

Pragmatics studies a set of interrelated phenomena and concepts that explain how context influences language interpretation. In this section, we introduce \textbf{what} pragmatics is, beginning with those core concepts, we then summarize the pragmatic phenomena examined in the surveyed works.

\subsection{Pragmatic Phenomena in Linguistics}
We summarize the following fundamental pragmatic phenomena from linguistics \cite[][\emph{inter alia}]{levinson1983pragmatics, yule1996pragmatics, birner2012introduction}: 

\textbf{Context.}
Context is the foundation of interpretating pragmatics, encompassing both linguistic context (e.g., surrounding clauses or sentences) and extra-linguistic context, including temporal, spatial, and social factors \cite{huang2017introduction}.

\textbf{Deixis.}
In particular, deictic expressions rely heavily on the situational context of an utterance for their interpretation. These expressions include personal deixis (e.g., “I”, “you”), spatial deixis (e.g., “this”, “that”), and temporal deixis (e.g., “now”, “then”), and are interpreted based on factors such as the speaker’s location, the time of the utterance, and the prior discourse \cite{levinson2006deixis, stapleton2017deixis, levinson1983pragmatics}.

\textbf{Implicature.} 
\citet{Grice1975}'s notion of implicature describes how speakers imply additional meanings without stating them explicitly. For example, if Ann asks, ``Do you sell paste?'' and Bill replies, ``I sell rubber cement'', the implication is that Bill does not sell paste \cite{Hirschberg1985ATO}. Implicature results from enriching the context 
with additional assumptions to comply with
the cooperative principle and the conversational maxims: truthfulness (Maxim of Quality), informativeness (Maxim of Quantity), relevance (Maxim of Relevance), and clarity (Maxim of Manner) \citep{Grice1975}.

\textbf{Presupposition.} Presuppositions are the implicit assumptions that must hold true for an utterance to be meaningful and understood \cite{stalnaker1977pragmatic}. For instance, the sentence ``The Queen of England is bald'' presupposes that England has a unique Queen, even though this fact is not directly stated \cite{stalnaker1977pragmatic}.

\textbf{Speech Acts and Intent Recognition.}
Understanding speech acts is critical for understanding speaker intent. 
A speech act is an utterance that not only conveys information but also performs an action \cite{austin1975things}.
Each utterance can be classified as performing an act that fulfills a communicative purpose, such as such as assertion, suggestion or description. For example, the utterance “Can you open the window?” may literally ask about the addressee’s ability or, depending on the context, function as an indirect request.

\textbf{Discourse and Coherence.}
In linguistics, discourse refers to language use beyond sentence level \cite{Guardado+2018+70+78}. Discourse structure is analyzed by examining how sentences and larger text units are interconnected through coherence relations such as elaboration, explanation and contrast. Some theoretical approaches offer hierarchical models that map these relations to reveal the underlying structure of texts (for an overview of Rhetorical Structure Theory, see \citealp{mann1988rhetorical}), while other frameworks provide a formal semantic account that integrates these coherence relations into dynamic discourse representation (for insights into Segmented Discourse Representation Theory, see \citealp{asher2003logics}).

\textbf{Social Pragmatics.}
Social pragmatics broadens traditional theories by exploring how social factors like culture, power, gender, and interpersonal dynamics influence language use. \citet{brown1987politeness} introduced the concept of “face” (self-esteem), showing how politeness strategies manage relationships and preserve social harmony. Gender differences in communication, as examined by \citet{tannen1990} and \citet{eckert1992think}, highlight contrasting conversational styles between men and women. Cultural variations further complicate communication, as \citet{scollon2011intercultural} demonstrate that politeness norms differ across cultures. In the digital era, social pragmatics also informs computer-mediated communication, where cues like emoticons convey emotions and tone \cite{herring2013introduction}. 

\subsection{Pragmatic Phenomena in the Survey} 
We next examine the pragmatic phenomena addressed in NLP, which were studied in the surveyed works. An overview of the survey methods and inclusivity is presented in Appendix \S\ref{sec:appendix_overview}. 
Since pragmatic concepts are interconnected, there are overlaps between categories. Therefore, we focus on the key aspects that each paper addresses. Figure \ref{fig:pheno} gives an overview of the phenomena and papers.

\textbf{Context and Deixis.} Context and deixis in NLP evaluation tasks usually assess a model's ability to interpret inputs based on the situational or linguistic context. 
The examination of the context and deixis is the basis of evaluating the models' pragmatic ability as the models rely on the context to respond, similar to the human response process \cite{Greasley2016,van-dijk-etal-2023-large,ma-etal-2024-potential}. Datasets or frameworks like \citet{min-etal-2020-ambigqa,sravanthi-etal-2024-pub,li2023diplomat,qi-etal-2023-pragmaticqa,sileo-etal-2022-pragmatics, hu-etal-2023-fine, shaikh-etal-2023-modeling,park-etal-2024-multiprageval,park2024pragmatic,guesswhat_game} explicitly address deixis as one of their pragmatic phenomena, requiring models to resolve references in context. In addition, \citet{li2023diplomat} and \citet{qi-etal-2023-pragmaticqa} incorporate context-dependent reasoning, 
requiring models to interpret utterances within multi-turn dialogues.
\citet{monroe-etal-2017-colors,monroe-etal-2018-generating} explore language use in color reference tasks, demonstrating how context influences referential choices. \citet{westera-etal-2020-ted} and \citet{zeyrek-etal-2018-multilingual} also emphasize the role of context in discussion structure and question-under-discussion frameworks. 
These datasets highlight the role of context in grounding language understanding. 

\textbf{Implicature and Presupposition.} 
A recurring theme across many papers is the study of implicature, i.e., testing the model's inference ability based on the textual input across literal meaning. It explores how meaning is implied rather than explicitly stated, requiring models to infer intentions and assumptions beyond literal interpretations. Different types of implicatures have been studied: \citet{qi-etal-2023-pragmaticqa}, \citet{hu-etal-2023-fine} and \citet{halat-atlamaz-2024-implicatr} focus on conversational implicatures, testing models' ability to infer pragmatic meanings in dialogue. \citet{nizamani-etal-2024-siga} investigate scalar implicatures, particularly with gradable adjectives. \citet{zheng-etal-2021-grice} use a grammar-based approach to generate dialogues with intricate implicatures. \citet{Cong2024} looks into LLMs' understanding of manner implicature, a pragmatic inference triggered by a violation of \citet{Grice1975}'s manner maxim. Presupposition is another aspect of implicature and influences the model's inference ability on the implications. \citet{louis-etal-2020-id,damgaard-etal-2021-ill,muller-plank-2024-indirectqa} focus on indirectness in dialogue. \citet{sap-etal-2020-social} work on implicatures posted in social media and study social biases. \citet{kameswari-etal-2020-enhancing} use pragmatic presupposition to enhance bias detection in political news. Moreover, additional works \cite{jeretic-etal-2020-natural,koyano-etal-2022-annotating,srikanth-etal-2024-pregnant,kimyeeun-etal-2024-developing} use pragmatic inferences to understand meanings of the given input.

\textbf{Speech Acts and Intent Recognition.} Speech acts have been explicitly studied in the surveyed works, usually in cases where there are requests, commands, or promises, given to the model and test the model's intent recognition. \citet{li2023diplomat} include tasks for pragmatic identification and reasoning, requiring models to recognize speech acts in multi-turn dialogues. \citet{reinig-etal-2024-politics} provide a fine-grained annotation of speech acts in parliamentary debates. \citet{braga-etal-2006-progmatica} link prosodic patterns to speech acts and discourse events. A few papers have explicitly different ``speakers'' involved in the evaluation frameworks and assess how well the models play as the ``speakers'' in the dialogues, performing the commended task. \citet{khani-etal-2018-planning} explore how players infer intentions and generate pragmatic messages in collaborative tasks, focusing on the rational speech act between the two agents (the ``speaker'' and the ``listener''). \citet{shaikh-etal-2023-modeling} involve a multi-turn collaborative two-player game based on the codenames. \citet{ollagnier-2024-cyberagressionado} also incorporates speech act-like annotations, such as ``attack'' and ``defend'', to capture the pragmatic roles of messages in multiparty chats. Both \citet{zhang-etal-2018-personalizing} and \citet{welleck-etal-2019-dialogue} talk about the personas, who are the agents making requests, in dialogue and work on the consistency between persona and dialogue. 

\textbf{Discourse and Coherence.} 
Discourse and coherence have been studied in NLP with a focus on parsing text into discourse relations \citep[e.g.,][]{miltsakaki-etal-2004-penn,prasad-etal-2008-penn,prasad-etal-2018-discourse,miao-etal-2024-discursive}, on discourse markers \citep[e.g.,][]{pandia-etal-2021-pragmatic,sadlier-brown-etal-2024-useful}, and on modeling dimensions of coherence \citep[e.g.,][]{lai-tetreault-2018-discourse,wu-etal-2023-multi-task}. 
In addition, discourse structures play a crucial role in dialogue and communication. 
\citet{westera-etal-2020-ted} and \citet{zeyrek-etal-2018-multilingual} investigate the role of evoked questions in discourse structure using TED talks. \citet{asher-etal-2016-discourse} explore discourse structure in multi-party dialogues and \citet{reinig-etal-2024-politics} study multi-party dialogues in German parliament. 
Moreover, discourse modes and aspect selection are influenced by contextual and pragmatic factors, connecting coherence to pragmatic interpretation \citep{mavridou-etal-2015-linking}. For a comprehensive overview of situation types and aspects, see \citet{friedrich-etal-2023-kind}.

\textbf{Social Pragmatics.} Social pragmatics explores how language use is shaped by social norms, power dynamics, and cultural contexts, including the roles of agents in communication. \citet{ollagnier-2024-cyberagressionado} introduces a tagset for annotating discursive roles in cyberbullying, while \citet{sap-etal-2019-social} and \citet{sap-etal-2020-social} provide benchmarks for reasoning about social norms, biases, and power dynamics. \citet{shaikh-etal-2023-modeling} highlight the role of shared cultural knowledge in communication through a collaborative word game, and \citet{zhang-etal-2018-personalizing} and \citet{welleck-etal-2019-dialogue} emphasize using personal and social context to improve dialogue systems. Additionally, \citet{yang-etal-2021-predicting} examine how pragmatic features vary between neurotypical and neurodiverse speakers, particularly adults with autism. These datasets collectively demonstrate the intersection of social pragmatics with other phenomena like discourse, implicature, and speech acts.

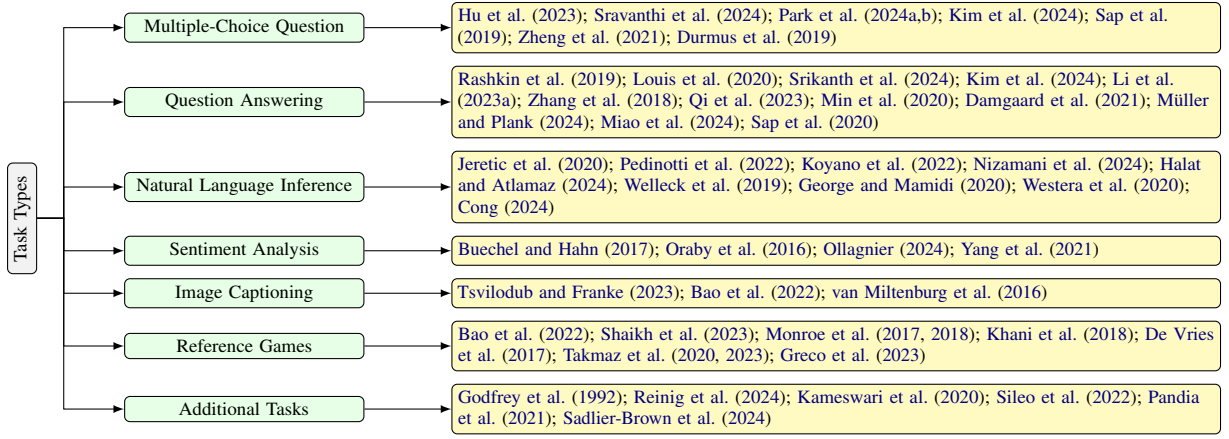
\begin{figure*}[htbp]
\scriptsize
%
\tikzset{
    basic/.style  = {draw, text width=1.3cm, align=center, 
    rectangle, fill=green!10},
    root/.style   = {basic, rounded corners=2pt, thin, align=center, fill=gray!10, rotate=0},
    tnode/.style = {basic, thin, rounded corners=2pt, align=left, fill=yellow!30, text width=10cm, anchor=center},
    xnode/.style = {basic, thin, rounded corners=2pt, align=center, fill=green!10,text width=3cm, anchor=center}, 
    edge from parent/.style={draw=black, edge from parent fork right}
}
\begin{forest} for tree={
    grow=east,
    growth parent anchor=east,
    parent anchor=east,
    child anchor=west,
    edge path={\noexpand\path[\forestoption{edge}, ->, >={latex}] 
        (!u.parent anchor) -- +(10pt,0) |- (.child anchor) \forestoption{edge label};},
}
[Task Types, root, l sep=11.5mm, rotate=90, child anchor=north, parent anchor=south, anchor=center, 
    [Additional Tasks, xnode,  l sep=11.5mm,
        [\citet{Godfrey1992,reinig-etal-2024-politics,kameswari-etal-2020-enhancing,sileo-etal-2022-pragmatics,pandia-etal-2021-pragmatic,sadlier-brown-etal-2024-useful}, tnode] 
        ]
    [Reference Games, xnode,  l sep=11.5mm,
        [\citet{bao-etal-2022-learning,shaikh-etal-2023-modeling,monroe-etal-2017-colors,monroe-etal-2018-generating,khani-etal-2018-planning,guesswhat_game,takmaz-etal-2020-refer,takmaz-etal-2023-speaking,Greco2023}, tnode] 
        ]
    [Image Captioning, xnode,  l sep=11.5mm,
        [\citet{tsvilodub-franke-2023-evaluating,bao-etal-2022-learning,van-miltenburg-etal-2016-pragmatic}, tnode] 
        ]
    [Sentiment Analysis, xnode,  l sep=11.5mm,
        [\citet{buechel-hahn-2017-emobank,oraby-etal-2016-creating,ollagnier-2024-cyberagressionado,yang-etal-2021-predicting}, tnode] 
        ]
    [Natural Language Inference, xnode,  l sep=11.5mm,
        [\citet{jeretic-etal-2020-natural,pedinotti-etal-2022-pragmatic,koyano-etal-2022-annotating,nizamani-etal-2024-siga, halat-atlamaz-2024-implicatr,welleck-etal-2019-dialogue,George2019ConversationalII,westera-etal-2020-ted,Cong2024}, tnode]
        ]
    [Question Answering, xnode,  l sep=11.5mm,
        [ \citet{rashkin-etal-2019-towards,louis-etal-2020-id,srikanth-etal-2024-pregnant, kimyeeun-etal-2024-developing,li2023diplomat,zhang-etal-2018-personalizing,qi-etal-2023-pragmaticqa,min-etal-2020-ambigqa,damgaard-etal-2021-ill,muller-plank-2024-indirectqa,miao-etal-2024-discursive,sap-etal-2020-social}, tnode]
        ] 
    [Multiple-Choice Question, xnode,  l sep=11.5mm,
        [\citet{hu-etal-2023-fine, sravanthi-etal-2024-pub, park2024pragmatic,park-etal-2024-multiprageval,kimyeeun-etal-2024-developing,sap-etal-2019-social, zheng-etal-2021-grice,durmus-etal-2019-role}, tnode]
        ] 
        ]
\end{forest}
\caption{A taxonomy of task types.}
\label{fig:task_type}
\end{figure*}

\section{Task Types}
\label{sec:task_types}
Understanding pragmatic language use requires evaluating models on a diverse set of tasks that capture various communicative functions and reasoning processes. 
We then summarize \textbf{how} the surveyed works have evaluated pragmatic phenomena, organizing tasks according to general NLP task classifications. 
They cover a broad spectrum, from structured multiple-choice question answering to open-ended question answering and reasoning, dialogue modeling, and multimodal challenges. A tree graph is shown in Figure \ref{fig:task_type}.

\textbf{Multiple-Choice Question (MCQ) Setup.} MCQs present a scenario or question with predefined answer options, prompting selection of one or more choices. 
This is commonly used to evaluate LLMs by asking them to choose the correct interpretation of an utterance \cite{hu-etal-2023-fine, sravanthi-etal-2024-pub, park2024pragmatic, park-etal-2024-multiprageval, kimyeeun-etal-2024-developing}. It also benchmarks pretrained QA models \cite{sap-etal-2019-social, zheng-etal-2021-grice}. Answers can include scalable items or Likert scales \cite{durmus-etal-2019-role}. MCQs simplify decision-making and improve response accuracy, making them effective for both human studies and LLM evaluations \cite{groves2011, ma-etal-2024-potential}.

\textbf{Question Answering (QA).} QA, including Conversational QA, models information-seeking dialogues by mapping questions to answers, enabling interactive communication. It evaluates reasoning about unspoken intent, a key aspect of human communication \cite{Grice1975}. QA tasks typically involve QA pairs \cite{rashkin-etal-2019-towards,louis-etal-2020-id,srikanth-etal-2024-pregnant, kimyeeun-etal-2024-developing,min-etal-2020-ambigqa} or multi-turn dialogues \cite{li2023diplomat, zhang-etal-2018-personalizing}. Some datasets include direct or literal answers to assess models' reasoning abilities \cite{qi-etal-2023-pragmaticqa,miao-etal-2024-discursive}.

\textbf{Natural Language Inference (NLI).} NLI usually includes a premise and a hypothesis to determine whether the hypothesis is true (entailment), false (contradiction), or undetermined (neutral) given the premise \cite{storks2020recent}. Pragmatic evaluations often include sentence pairs with certain different pragmatic particles or words \cite{jeretic-etal-2020-natural,pedinotti-etal-2022-pragmatic,koyano-etal-2022-annotating,nizamani-etal-2024-siga, halat-atlamaz-2024-implicatr} or sentences from dialogues \cite{welleck-etal-2019-dialogue}, testing whether models can correctly select certain pragmatic (un)related sentences. 
There are also datasets with NLI-like features looking at the relations of given sentences \cite{George2019ConversationalII,westera-etal-2020-ted,Cong2024}. 

\textbf{Sentiment Analysis.} Sentiment analysis in pragmatics, unlike analysis which focuses on polarity \cite{gong-etal-2024-mapping}, extends traditional sentiment classification by incorporating communicative intent, discourse roles, and social interaction dynamics. Datasets in this domain are designed to capture these nuances through fine-grained annotations of conversational exchanges, in topics such as emotion \cite{buechel-hahn-2017-emobank}, sarcasm \cite{oraby-etal-2016-creating}, hatefulness \cite{ollagnier-2024-cyberagressionado}, or medical applications for people with Autism Spectrum Disorder (ASD, \citealt{yang-etal-2021-predicting}). 

\textbf{Image Captioning.} Image captioning involves generating descriptions that convey the meaning of a given image. It is a crucial task for studying human reasoning about alternatives and efficient consideration of context \cite{fried-etal-2023-pragmatics}. Unlike other NLP tasks, it is inherently grounded in visual content. Datasets in this area are often designed to investigate pragmatic phenomena, such as contrastive captions and speaker-listener dynamics \cite{tsvilodub-franke-2023-evaluating, bao-etal-2022-learning}. Some explores negation-based descriptions to enhance pragmatic reasoning \cite{van-miltenburg-etal-2016-pragmatic}. These datasets emphasize the interaction between visual context and textual interpretation.

\textbf{Reference Games.} 
According to \citet{fried-etal-2023-pragmatics}, reference games involve a speaker describing a target referent from a shared set of images, objects, or abstract illustrations to a listener, who must then identify the target. 
This setup is often used to model pragmatic reasoning, such as in tasks requiring context-based contrastive descriptions or multi-turn interactions \cite{bao-etal-2022-learning, shaikh-etal-2023-modeling}. Variants include adaptations of human games to study sequential language use \cite{monroe-etal-2017-colors, khani-etal-2018-planning} as well as visual language games in guessing given objects given a visual context \cite{guesswhat_game,takmaz-etal-2020-refer,takmaz-etal-2023-speaking,Greco2023}, examining the ability to understand specific referential terms. These datasets emphasize context-dependent interpretation and collaborative communication. 

\textbf{Additional Tasks.} 
There are additional task types designed specifically for certain pragmatic features, including speech-based tasks \cite{Godfrey1992,reinig-etal-2024-politics} for speech act studies, 
bias detection  \cite{kameswari-etal-2020-enhancing} for assessing social norms, cloze-style tasks for discourse markers \cite{pandia-etal-2021-pragmatic,sadlier-brown-etal-2024-useful},  and various natural understanding tasks curated for a pragmatics-centered evaluation framework \cite{sileo-etal-2022-pragmatics}.

\section{Dataset Construction}
\label{sec:collection}

In this section, we summarize methods and data sources used in previous studies for building a pragmatic dataset. 
Discussing the dataset construction process is crucial for understanding the scope of current pragmatic datasets in terms of domains and source data, thereby gaining insights about their limitations and special features.  
We categorize dataset-building approaches for evaluating pragmatic phenomena in more depth, suggesting two groups: a \textbf{bottom-up approach}, where source data is first collected and annotation for certain pragmatic phenomena is then performed; and a \textbf{top-down approach} which first determines labeling among seed data (e.g., scalar pairs), and then expands seed data to large units, such as sentences, to build up the whole dataset. 

\subsection{Bottom-up Approach} 
We discuss this approach in twofold: collecting source data and annotating pragmatic phenomena. 

\textbf{Source Data.}
In general, there are three types of source data that current pragmatic datasets are built upon: (1) databases, such as web pages and interviews, (2) data collected directly from humans and (3) existing datasets. 

There is no one specific \textbf{database} for building a pragmatic dataset.
Works focusing on dialogue or speech acts use interviews or television programs \citep{braga-etal-2006-progmatica, George2019ConversationalII} as source data.
Common domains for sourcing pragmatic data include open domain \citep{qi-etal-2023-pragmaticqa, wang-etal-2020-dusql}, debate \citep{durmus-etal-2019-role, oraby-etal-2016-creating}, politics \citep{reinig-etal-2024-politics}, social media \citep{ollagnier-2024-cyberagressionado,sap-etal-2020-social} and law \cite{kimyeeun-etal-2024-developing}.
Fewer works \textbf{collect data from humans} due to difficulties in recruiting human participants.
For instance, \citet{yang-etal-2021-predicting} collected spoken language data from adults with high-functioning ASD to examine their pragmatic features.
Other works collect data in a game setup, such as reference games, focusing on aspects such as context usage \cite{khani-etal-2018-planning} and pragmatic inference \citep{shaikh-etal-2023-modeling, monroe-etal-2017-colors}.
The majority of datasets are built upon \textbf{existing datasets}.
They are either a unified benchmark extended on an existing pragmatic dataset \citep{sravanthi-etal-2024-pub,sileo-etal-2022-pragmatics, park-etal-2024-multiprageval, guo-etal-2021-chase},
or built by annotating non-pragmatic datasets \citep{li2023diplomat,bao-etal-2022-learning, welleck-etal-2019-dialogue,kameswari-etal-2020-enhancing,ollagnier-2024-cyberagressionado, van-miltenburg-2017-pragmatic}.
Conversely, another line of research uses existing datasets not originally related to pragmatics to explore pragmatic phenomena. 
For example, \citet{ollagnier-2024-cyberagressionado} adds intent labels to analyze implicature in hate speech.

\textbf{Annotation.}
Typical annotation methods involve crowd-sourced annotations \citep{qi-etal-2023-pragmaticqa,li2023diplomat, shaikh-etal-2023-modeling} and expert annotations \citep{jeretic-etal-2020-natural,sravanthi-etal-2024-pub, hu-etal-2023-fine, park2024pragmatic}.
LLMs can also be used to assist annotation. 
For instance, to create a QA dataset featuring implicature, \citet{srikanth-etal-2024-pregnant} utilize GPT-3.5 to consolidate a list of assumptions and sub-questions annotated by experts.
The model needs to either turn a sub-question into a declarative sentence to create an implicature inference, or identify the assumptions and implicature made in the sub-question.
Though speeding up the annotation process and assuring data format, in the post analysis, they find using LLMs to directly generate implicature inference is not reliable.
Besides, \citet{Cong2024} employs ChatGPT in generating synthetic stimuli for implicature sentences, with human raters reviewing after the automatic generation process, with the setup of Likert scaling scoring. 
These hybrid approaches can facilitate the efficient processing of large datasets while maintaining a high level of accuracy through human participation.

\subsection{Top-down Approach} 
Unlike the bottom-up approach, where labels are created after data collection and usually provided by annotators, the labeling process in the top-down approach is motivated by linguistic theories and can be done automatically \cite{halat-atlamaz-2024-implicatr, nizamani-etal-2024-siga, koyano-etal-2022-annotating, sap-etal-2019-social}. For instance, to build an NLI dataset featuring implicature, \citet{halat-atlamaz-2024-implicatr} first curate a list of scalar pairs from various linguistic categories. These scalar pairs encode logical relationships naturally, e.g., \textit{<some, all>}. Then they employ GPT-4 to generate sentences based on human-created examples as demonstrations (\textit{I read some of the books.}~v.s.~\textit{I read all of the books.}). Similarly, to build a pragmatic NLI dataset, \citet{nizamani-etal-2024-siga} collect sentences containing scalar adjectives separated by ``\textit{but not}'' (e.g., \textit{good but not great}), motivated by the observation that implicatures can be explicitly reinforced \cite{Hirschberg1985ATO}. To create premise-hypothesis pairs, they split the sentence into two parts, each containing one scalar adjective. For instance, the original sentence: ``\textit{These shows were good, but not great.}'' can be split into: \textit{``These shows were good.}'' and \textit{``These shows were great.}''
Manual modification ensures sentence independence and grammatical correctness. 

\section{Evaluation}
\label{sec:eval}

We analyze the evaluation of pragmatics in the previous works by summarizing the metrics and methods employed, highlighting their advancements and limitations in capturing pragmatic phenomena.

\textbf{Evaluation on Model Performance.} 
Currently, automatic metrics dominate the evaluation landscape, with evaluation methods primarily focusing on measuring how well models align with annotated labels or golden data. While this approach is effective for tasks with well-defined categories, it often fails to capture the complexity and subtlety of pragmatic phenomena. For instance, in classification tasks, where average F1 score is widely used, many classification tasks produce binary or categorical labels that reflect only a narrow aspect of pragmatics, such as speech acts \cite{reinig-etal-2024-politics} or NLI \cite{nizamani-etal-2024-siga}, without addressing broader contextual or interactive dimensions. Similarly, emotion prediction \cite{buechel-hahn-2017-emobank} relies on labels (e.g., “joy”, “anger”) which misses nuanced constructs such as empathy or politeness, and in some applications, the task diverges entirely from pragmatics and uses a binary label for downstream application results. 
Generation tasks often rely on ROUGE, BLEU, and BERTScore, as seen in cross-cultural pragmatic inference \cite{shaikh-etal-2023-modeling} and social pragmatics \cite{sap-etal-2020-social}. Other metrics include factuality \cite{qi-etal-2023-pragmaticqa}, accuracy \cite{li2023diplomat, kimyeeun-etal-2024-developing}, and task-specific metrics tailored to individual research goals \cite{li-etal-2023-pragmatic}. While these metrics provide valuable insights into model performance, they often fall short in evaluating the broader, interconnected nature of pragmatics across tasks.

\textbf{Beyond Automatic Evaluation.} 
Although automatic metrics like F1 score and ROUGE are useful for efficiency and standardization, they can fall short in capturing more nuanced pragmatic aspects, such as politeness. To this end, human evaluation is still needed for more comprehensive evaluation.
For instance, in \citet{yang-etal-2021-predicting}, trained human annotators were asked to rate the scale of politeness and uncertainty in the spoken language data from adults with high-functioning ASD from $1$ to $3$.
These investigated speech features cannot be easily measured by automatic evaluation metrics. 
Likewise, \citet{rashkin-etal-2019-towards} ran a crowd-sourcing human evaluation on MTurk, where annotators determined whether model-generated responses show empathy. 
Hence, dialogue models are evaluated not only for its general text generation ability (capturable by automatic metrics), but also more nuanced aspects in human communication, such as emotions.
Such a hybrid framework provides a more complete picture of model performance, especially in terms of their capabilities of language uses beyond understanding surface meanings.  

\textbf{}
We believe it is a promising direction to integrate both automatic and manual evaluation methods that better capture the nuanced and multi-dimensional nature of pragmatic phenomena. 
Furthermore, there is a growing need for holistic evaluation methodologies that assess a model’s ability to handle multiple pragmatic phenomena simultaneously, such as combining emotion detection, politeness, and social reasoning in complex, real-world scenarios \cite{wu-etal-2024-rethinking, van-dijk-etal-2023-large, sileo-etal-2022-pragmatics}. These advancements would enable a more robust understanding of how well models capture the full spectrum of pragmatic behavior. 
We further discuss this point with an extension to an outlook towards the high-level evaluation frameworks in \S\ref{sec:discussion_eval}.

\section{Opportunities and Challenges}
\label{sec:discussion}
We now discuss the opportunities and challenges identified in the surveyed resources, with a focus on the rapid development of LLMs. We focus on generalizability, the high-level evaluation outlook, alignment, as well as the potential of LLMs to advance research in linguistic pragmatics both within and beyond existing resources.

\subsection{Gaps and Chances in Generalizability}
\label{sec:generalizability}
Based on the surveyed papers, we identify several gaps in the generalizability of current research.

\textbf{English-Centric Bias.} Current pragmatic research is dominated by English-focused datasets, limiting cross-linguistic and cultural generalizability. Out of all $57$ surveyed papers, only $11$ (i.e., $19$\%) resources include non-English languages.  
Though there are a few non-English resources, such as German \cite{reinig-etal-2024-politics} and Korean \cite{park2024pragmatic}, they focus on regional contexts, lacking critical perspectives for understanding cross-linguistic and cross-cultural pragmatics.

\textbf{Human/Demographic Bias.} NLP frameworks that acknowledge that in certain tasks human labels and judgments may be inherently subjective, and display a lot of variation between subjects (\citealp[see e.g.,][]{plank-2022-problem,Cabitza_Campagner_Basile_2023,kern-etal-2023-annotation}). In addition, recently, more and more works have focused on simulating human behaviours and opinions including social demographic profiles of real human participants (\citealp[e.g.,][]{argyle-etal-2023}). The demographics aspect is in general an important factor in the data collection phase in terms of human label variation, and should be considered while conducting LLM evaluations, which we see as a gap in current works. 

\textbf{Data Type Diversity.} Most datasets remain largely uni-modal, focusing on either text or speech, with few exploring interactions across modalities and genres. For instance, SWITCHBOARD \cite{Godfrey1992}, while valuable for spoken dialogue, lacks visual data, such as gestures or facial expressions, that are essential for modeling phenomena like deixis, implicature, or politeness. A multimodal dataset combining speech, text, and video could address complex contexts like sarcasm or ambiguity resolution, aligning more closely with real-world communication and advancing pragmatic competence evaluation.

\textbf{Limitations in Task Types.} Pragmatic tasks typically emphasize one or a small set of phenomena, such as speech acts or emotion detection, without capturing the broader interplay between multiple pragmatic dimensions. For example, datasets like SWITCHBOARD \cite{Godfrey1992} focus on spoken dialogue and discourse markers, while the Lexical Markup Framework \cite{braga-etal-2006-progmatica} targets lexical standardization. These tasks are often designed in isolation, making it difficult to evaluate a model's holistic pragmatic competence. Similarly, political discourse datasets like \citet{reinig-etal-2024-politics} analyze specific speech acts but do not integrate additional pragmatic phenomena, such as implicature or politeness. This fragmented approach limits our ability to assess how well models handle the complex, multifaceted nature of pragmatics in real-world scenarios.

\textbf{Towards Generalizability.} Addressing these gaps requires the creation of \textbf{multilingual, multimodal datasets} that incorporate visual, textual, and spoken interactions. Leveraging hybrid approaches that combine synthetic LLM-generated data with human validation could provide scalable and diverse benchmarks. In addition, pragmatic task designs should move beyond uni-modal classification and generation to explore holistic, real-world applications that evaluate models' ability to integrate multiple pragmatic phenomena, such as emotion detection, politeness, and implicature resolution, within the same task framework. These advancements will enable a more comprehensive and inclusive understanding of pragmatic competence in computational models.

\subsection{Fine-Grained Evaluation of Pragmatics}
\label{sec:discussion_eval}

Despite the growing number of datasets and benchmarks designed to evaluate specific aspects of pragmatic abilities in NLP models, particularly LLMs, there remains a critical need for a comprehensive, high-level framework for fine-grained pragmatic evaluation. 
Below, we outline key gaps in current evaluation practices and propose recommendations for advancing fine-grained pragmatic evaluation.

\textbf{Metrics for Pragmatic Alignment.}
Current evaluation metrics, largely based on traditional NLP tasks, fail to capture the nuances of how well LLMs handle context and social norms. More dynamic and context-sensitive measures, including human-in-the-loop assessments, are needed to assess how well models align with pragmatic expectations. For instance, metrics from psycholinguistics and psychometrics \cite{shu-etal-2024-dont} are useful to assess the LLM responses. 

\textbf{In-depth Analysis of LLM Outputs.} In \S\ref{sec:task_types}, we drew a diverse landscape of the NLP tasks that the datasets cover. However, most studies treat pragmatics as a classification or generation task, or one of the aspects in their task settings, without closely examining how models arrive at their responses. 
A more in-depth, qualitative approach—potentially incorporating human feedback and error analysis—could offer deeper insights into where models succeed or fail in pragmatic reasoning. This is particularly relevant for tasks requiring subtle social inferences, implicature resolution, or sensitivity to power dynamics in conversation. We need better measures for evaluating the thought competence in LLMs \cite{van-dijk-etal-2023-large}. 

\textbf{Expanding Data Resources.} Unlike syntactic or semantic tasks, pragmatic evaluation relies on rich, contextually grounded datasets, which are difficult to construct at scale. While crowdsourcing and expert annotation are viable solutions, they are resource-intensive. A possible direction is to leverage LLMs themselves to generate controlled datasets, though augmented by human validation \cite{long-etal-2024-llms}. For instance, recent works \cite{Cong2024,srikanth-etal-2024-pregnant} use GPT to generate synthetic stimuli, with validations from human evaluators. 

\subsection{Pragmatics Aids LLM Alignment}
Recent research emphasizes incorporating linguistic insights into LLM development \cite{opitz2024natural,brunato-2025-learning}. We also advocate insights from pragmatics for LLM alignment.
The alignment of LLMs refers to their ability to generate responses that are not only factually correct but also contextually appropriate and socially coherent \cite{shen2023largelanguagemodelalignment}. For instance, speech act theory \cite{Searle_1969} highlights that meaning emerges from interaction, not isolated sentences—aligning with the communicative goals of LLMs.

\textbf{Pragmatics in Instruction and Alignment.} LLMs rely on prompts to infer user intent, making pragmatics essential for refining instruction-following. Studies have used pragmatic frameworks to evaluate and adjust model responses (\citealp[e.g.,][]{Ruis2022TheGO,sravanthi-etal-2024-pub,yerukola-etal-2024-pope}), yet they lack robustness in pragmatic inference, failing to recognize indirect requests or subtle shifts in social meaning \cite{Lee_Daniel_2024}. 
By systematically incorporating pragmatic constraints and examples, LLMs can better interpret ambiguous user inputs and produce responses that align with conversational expectations \cite{vaduguru2024generating}.

\textbf{Human-Computer Interaction.} Pragmatics can improve human-AI dialogue and collaboration by ensuring contextually appropriate and socially aware responses. However, chatbot alignment remains limited, especially in handling ambiguity and implicit meaning \cite{martinenghi-etal-2024-llms}. Incorporating pragmatic reasoning can enhance user satisfaction by making interactions more intuitive and responsive to nuanced human intentions, contributing to the continuous advancement of language models \cite{vaduguru2024generating}. 

\textbf{Multi-Agent and Collaborative AI.}
Pragmatics is crucial in multi-agent systems, where agents must infer intent from indirect cues \cite{li2024pacepragmaticagentenhancing}. While a single LLM can role-play different characters with a well-crafted prompt, communication degrades when multiple agents interact, due to LLMs’ lack of true belief-state reasoning \cite{zhou2024reallifejustfantasy}. Addressing this requires models that better handle information asymmetry and implicit intent in multi-agent dialogue.

\subsection{LLMs Enhance Research in Pragmatics}
Finally, we return to the research in linguistic pragmatics. Pragmatics is a core area of linguistics that has increasingly embraced experimental methods over the past two decades \cite{sauerland2025pragmatics}. Recent advances in LLMs offer new opportunities to advance pragmatic research.

\textbf{Stimulus and Material Design.} Experimental pragmatics relies heavily on carefully designed stimuli to test specific hypotheses \cite{schwarz2017experimental}. The recently evolving LLMs might be able to facilitate this by generating diverse and context-rich materials that mirror real-world communication. This can significantly streamline the experimental design process for pragmatic phenomena. 

\textbf{Supporting Data Annotation.} 
High-quality annotation is crucial for pragmatic research but is often costly and time-consuming. LLMs can support the annotation process by providing preliminary labels that human annotators can refine. This hybrid approach has shown promise in previous studies \cite{tan2024large} but requires careful implementation to ensure annotation quality and reliability \cite{nasution2024chatgpt, ronningstad2024gpt}. Combining LLM-assisted pre-annotation with thorough human review \cite{wang2024human} offers a more reliable  
approach to data annotation.

\textbf{Recommendations for Future Research in Linguistic Pragmatics.} Leveraging LLMs' potential in pragmatic research requires collaboration among computational linguists, cognitive scientists, and NLP practitioners. Researchers should explore integrating LLMs into experimental design, material generation, and task development to simulate complex conversational contexts. Rigorous human evaluation and qualitative studies are essential for reliability and interpretability.

\section{Conclusion}
This survey provides a comprehensive overview of existing resources for evaluating pragmatic understanding in NLP. By categorizing datasets according to the pragmatic phenomena they target and analyzing their task designs with data collection methods and how they are evaluated with the models, we highlight current trends and challenges in this field. Our findings underscore the need for more nuanced and contextually rich evaluation benchmarks, especially as LLMs continue to evolve. We hope to provide a valuable guide for researchers and practitioners advancing pragmatic reasoning in NLP, promoting systems with more human-like communication and interdisciplinary collaboration.

\section*{Limitations}

We identify the following limitations in the survey.

\textbf{Survey Sources.} Our survey predominantly encompasses literature within NLP, as the scope is to review the resources for evaluating pragmatics in NLP models. Therefore, we did not look extensively for resources in linguistic studies.
This might result in an omission of potential resources or enlightening evaluation methods, which could bring insights to current data collection processes and evaluations. We advocate for future surveys in reviewing the resources for pragmatic language understanding in linguistics.

\textbf{NLP Tasks.} Our survey is tailored towards tasks that feature applicability in NLP. 
As a result, intrinsic pragmatic tasks, such as discourse modeling are not discussed. 
Nevertheless, they could still be valuable to look into, especially in terms of gaining insights to current evaluation. 

\textbf{Multilinguality.} The scope of our survey is confined to literature published only in English. 
Although we endeavored to include works addressing multilinguality (albeit written in English), it is conceivable that pragmatic datasets created in other languages are not incorporated.
This language constraint might limit the inclusivity of diverse pragmatic phenomenons and methodologies that are unique to a specific language.

\textbf{Multimodality.} While our paper covers a few data points that go beyond the textual-based features, such as the image captioning tasks (\citealp[e.g.,][]{tsvilodub-franke-2023-evaluating}), the majority of the data is textual-based. This gap has also been discussed in our §\ref{sec:generalizability}. As we mainly focus on NLP venues, some resources could be not covered. Therefore, we encourage future work to explore pragmatic evaluation across diverse modalities.

\section*{Ethical Considerations}
In this paper, we examine the evaluation of pragmatics in NLP, particularly in the context of LLMs. Since pragmatic abilities are inherently human, discussing these abilities in LLMs could raise concerns about anthropomorphism, as noted in prior work on LLM evaluation \cite{ma-etal-2024-potential,ma2024algorithmic}. It is important to clarify that our discussion of pragmatics does not imply human-like cognition or consciousness in LLMs. Instead, we advocate to use human pragmatic features to build benchmarks, aiming to enhance their design, and to enable user-friendly and effective human-machine interactions.

Moreover, our evaluation focuses on the pragmatic aspects reflected ``in'' LLMs' textual outputs, rather than attributing beliefs or intentions to the models themselves. This consideration is consistent with recent discussions on LLM evaluation, which emphasize analyzing model outputs without ascribing human-like agency (\citealp[e.g.,][]{santurkar2023whose,rottger-etal-2024-political, durmus2024towards}). By maintaining this perspective, we aim to contribute to a nuanced understanding of LLMs' capabilities while avoiding the pitfalls of over-personalization.

\section*{Acknowledgements}
The authors acknowledge the use of ChatGPT exclusively to paraphrase and refine the text in the final manuscript. 
We thank the members of the MaiNLP lab from LMU Munich for their constructive feedback.  
YL and KJ are supported by the Deutsche Forschungsgemeinschaft (DFG, German Research Foundation), Project-ID 281511265 – SFB 1252 “Prominence in Language” in the project C06 at the University of Cologne. ZG is supported by the National Science Foundation via ARNI (The NSF AI Institute for Artificial and Natural Intelligence), under the Columbia 2025 Research Project (“Towards Safe, Robust, Interpretable Dialogue Agents for Democratized Medical Care''). YJL and BP are supported by ERC Consolidator Grant DIALECT (101043235).

\bibliography{custom}
\appendix

\section{Overview of Surveyed Works}
\label{sec:appendix_overview}
To compile this survey, we conducted a comprehensive review of recent literature on evaluating pragmatics in NLP, with a specific focus on the datasets. We focused on identifying works that address these aspects, using the keywords ``pragmatics'' and ``datasets'' by primarily looking at papers publicized by 31.12.2024 at ACL Anthology\footnote{\url{https://aclanthology.org/}}.

This was chosen as the primary source since it is the main publication platform for the *CL community and contains the most relevant works on the evaluation of pragmatics in NLP models. After this initial search, we applied additional filtering criteria to refine the selection. Specifically, we focused on papers that 
\begin{itemize}
\item[-] introduced datasets explicitly designed for evaluating pragmatic phenomena in NLP, 
\item[-] discussed methodologies for assessing pragmatic abilities of language models, or 
\item[-] provided empirical evaluations of NLP models in related tasks to pragmatics. 
\end{itemize}

In addition, we conducted manual qualitative inspections to ensure the inclusion of relevant works that might not have been captured due to variations in terminology; we also included a few recent papers from other sources out of the ACL Anthology based on our expertise in the field. This resulted in a final section of 58 papers in the current version.

\section{Pragmatic Phenomena and their Corresponding Tasks}
\label{sec:phenomena-tasks}

In this section, we present the pragmatic phenomena and their corresponding tasks based on the survey works and show their mappings in Table \ref{tab:task2pheno}. We noticed that there are no 100 percent one-to-one mappings between the pragmatic phenomena and the tasks, i.e., a phenomenon could be evaluated in either task in reality. This shows the possibility of diverse formats to evaluate the pragmatics of LLMs, and calls for future development of finer-grained tasks.

\begin{table*}[b]
\footnotesize
\centering
\begin{tabular}{|p{2cm}|p{2cm}|p{2cm}|p{2cm}|p{2cm}|p{2cm}|p{2cm}|}
\hline
\textbf{Task Type} & \textbf{Context and Deixis} & \textbf{Implicature and Presupposition} & \textbf{Speech Acts and Intent Recognition} & \textbf{Discourse and Coherence} & \textbf{Social Pragmatics} \\
\hline
\textbf{Multiple-Choice Question} &
\citet{hu-etal-2023-fine, sravanthi-etal-2024-pub, park2024pragmatic, park-etal-2024-multiprageval} &
\citet{hu-etal-2023-fine,sravanthi-etal-2024-pub,zheng-etal-2021-grice, kimyeeun-etal-2024-developing} &
-- &
\citet{durmus-etal-2019-role} &
\citet{sap-etal-2019-social} \\
\hline
\textbf{Question Answering} &
\citet{li2023diplomat, qi-etal-2023-pragmaticqa, min-etal-2020-ambigqa, srikanth-etal-2024-pregnant, kimyeeun-etal-2024-developing, miao-etal-2024-discursive,sap-etal-2020-social} &
\citet{qi-etal-2023-pragmaticqa, louis-etal-2020-id, srikanth-etal-2024-pregnant, damgaard-etal-2021-ill, muller-plank-2024-indirectqa,kimyeeun-etal-2024-developing} &
\citet{li2023diplomat, zhang-etal-2018-personalizing,rashkin-etal-2019-towards} &
\citet{miao-etal-2024-discursive} &
\citet{yang-etal-2021-predicting,sap-etal-2020-social,zhang-etal-2018-personalizing} \\
\hline
\textbf{Natural Language Inference} &
\citet{westera-etal-2020-ted} &
\citet{jeretic-etal-2020-natural, halat-atlamaz-2024-implicatr, nizamani-etal-2024-siga, koyano-etal-2022-annotating, Cong2024,George2019ConversationalII,pedinotti-etal-2022-pragmatic} &
\citet{welleck-etal-2019-dialogue} &
\citet{westera-etal-2020-ted} &
\citet{welleck-etal-2019-dialogue} \\
\hline
\textbf{Sentiment Analysis} &
-- &
-- &
\citet{ollagnier-2024-cyberagressionado} &
\citet{yang-etal-2021-predicting} &
\citet{ollagnier-2024-cyberagressionado,yang-etal-2021-predicting,buechel-hahn-2017-emobank,oraby-etal-2016-creating}\\
\hline
\textbf{Image Captioning} &
\citet{tsvilodub-franke-2023-evaluating,bao-etal-2022-learning,van-miltenburg-etal-2016-pragmatic} &
-- &
-- &
-- &
-- \\
\hline
\textbf{Reference Games} &
\citet{guesswhat_game, monroe-etal-2017-colors,monroe-etal-2018-generating,takmaz-etal-2020-refer,takmaz-etal-2023-speaking,shaikh-etal-2023-modeling} &
-- &
\citet{shaikh-etal-2023-modeling,khani-etal-2018-planning,Greco2023} &
-- &
\citet{shaikh-etal-2023-modeling} \\
\hline
\textbf{Additional Tasks} &
\citet{sileo-etal-2022-pragmatics} &
\citet{kameswari-etal-2020-enhancing} &
\citet{Godfrey1992,reinig-etal-2024-politics} &
\citet{pandia-etal-2021-pragmatic, sadlier-brown-etal-2024-useful, reinig-etal-2024-politics} &
-- \\
\hline
\end{tabular}
\caption{Mapping from Pragmatic Phenomena (columns) to NLP Task Types (rows).}
\label{tab:task2pheno}
\end{table*}

\end{document}